\documentclass[10pt,twocolumn,letterpaper]{article}

\usepackage[pagenumbers]{iccv} 

\definecolor{iccvblue}{rgb}{0.21,0.49,0.74}
\usepackage[pagebackref,breaklinks,colorlinks,allcolors=iccvblue]{hyperref}
\usepackage{textcomp}
\usepackage{balance}

\def\paperID{*****} 
\def\confName{ICCV}
\def\confYear{2025}

\title{Capture, Canonicalize, Splat:\\Zero-Shot 3D Gaussian Avatars from Unstructured Phone Images}

\toggletrue{iccvfinal}

\author{Emanuel Garbin \and Guy Adam \and Oded Krams \and Zohar Barzelay \and Eran Guendelman \and Michael Schwarz \and Matteo Presutto \and Moran Vatelmacher \and Yigal Shenkman \and Eli Peker \and Itai Druker \and Uri Patish \and Yoav Blum \and Max Bluvstein \and Junxuan Li \and Rawal Khirodkar \and Shunsuke Saito\\[0.4em]
Meta
}

\begin{document}
\maketitle
\begin{abstract}
We present a novel, zero-shot pipeline for creating hyperrealistic, identity-preserving 3D avatars from a few unstructured phone images.
Existing methods face several challenges: single-view approaches suffer from geometric inconsistencies and hallucinations, degrading identity preservation, while models trained on synthetic data fail to capture high-frequency details like skin wrinkles and fine hair, limiting realism.
Our method introduces two key contributions: (1) a generative canonicalization module that processes multiple unstructured views into a standardized, consistent representation, and (2) a transformer-based model trained on a new, large-scale dataset of high-fidelity Gaussian splatting avatars derived from dome captures of real people.
This ``Capture, Canonicalize, Splat'' pipeline produces static quarter-body avatars with compelling realism and robust identity preservation from unstructured photos.
\end{abstract}
    
\begin{figure*}[ht]
  \centering
  \includegraphics{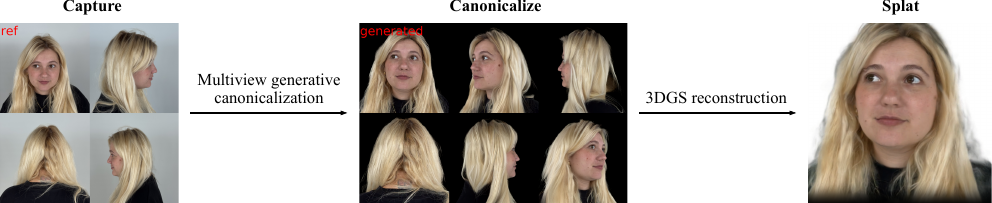}
  \caption{\textbf{Our ``Capture, Canonicalize, Splat'' pipeline.} From a few unstructured phone photos (left), our generative canonicalization module synthesizes a set of 3D-consistent, identity-preserving views with fixed camera parameters (middle). These views are then lifted by our transformer-based reconstruction model, trained on our high-fidelity dataset, into a hyperrealistic 3D Gaussian splatting avatar (right).}
  \label{fig:pipeline}
\end{figure*}

\section{Introduction}
\label{sec:intro}

The creation of photorealistic digital humans is a long-standing goal in computer vision and graphics, with applications ranging from virtual reality and telepresence to entertainment and digital fashion. The ultimate goal is to generate a faithful 3D representation of a specific individual, a "digital twin," from easily accessible inputs, such as a few photos taken with a smartphone.
However, creating such avatars from unstructured images presents significant challenges. Methods relying on a single input image often struggle with ambiguity, leading to geometric distortions and hallucinatory details, particularly for unseen parts of the person such as the back of the head \cite{lyu2024facelift, xu2024grm}. This geometric inconsistency invariably compromises the preservation of the subject's identity, a critical requirement for believable avatars. While multi-view reconstruction methods can produce more consistent geometry, they typically require calibrated camera setups, which are impractical for casual users.
Another major hurdle is achieving true realism. Many state-of-the-art generative models are trained on large-scale synthetic datasets such as RenderPeople\footnote{\url{https://renderpeople.com}}. Although these datasets provide diversity in identity, pose, and clothing, they often lack the high-frequency, person-specific details that define realism: subtle skin textures, fine wrinkles, and the intricate structure of hair. Models trained on such data can produce plausible humans but fail to capture the unique essence of a specific individual, often resulting in an "over-smoothed" or generic appearance.
To address these fundamental limitations, we propose a novel three-stage pipeline: ``Capture, Canonicalize, Splat.'' Our approach uniquely tackles the dual problems of identity preservation and hyper-realism.
First, instead of relying on a single, ambiguous view, our method takes a small set of unstructured phone images (\eg, front, back, and sides) as input. To handle the geometric inconsistencies inherent in such captures, we introduce a \textbf{generative canonicalization module}. This model processes the unstructured views and synthesizes a standardized, 3D-consistent set of multi-view images with known camera poses. By aggregating information from multiple inputs, it generates a complete and coherent representation of the person, significantly enhancing identity preservation.
Second, to achieve an unprecedented level of realism, we introduce a new training paradigm. We train our 3D lifting model on a novel large-scale dataset of \textbf{person-specific Gaussian splatting avatars}. These ground-truth 3D models are derived from high-quality dome captures of real individuals, retaining intricate details such as skin pores and fine hair strands. By training directly on these high-fidelity 3D representations, our model learns to generate avatars with exceptional person-specific realism.
Finally, the canonicalized views are lifted into a 3D representation by a transformer-based reconstruction model. Inspired by recent Large Reconstruction Models \cite{zhang2024gslrm}, it directly predicts a high-fidelity 3D Gaussian splatting \cite{kerbl20233d} representation, enabling high-quality rendering.
Our main contributions are:
\begin{itemize}
    \item A complete, zero-shot pipeline that generates hyperrealistic, identity-preserving static 3D avatars from a few unstructured phone images.
    \item A generative canonicalization module that normalizes unstructured multi-view inputs into a 3D-consistent representation, robustly preserving identity.
    \item A novel training methodology using a large-scale dataset of high-fidelity Gaussian splatting avatars to learn and reproduce fine, person-specific details.
\end{itemize}

\section{Related work}
\label{sec:related}

\paragraph{Single-image 3D human reconstruction.}
Creating a 3D human from a single image is a highly ill-posed problem. Early optimization-based methods relied on the fitting of parametric models such as SMPL \cite{smpl} to 2D joint detections \cite{bogo2016smplautomaticestimation3d}, while later learning-based approaches focused on direct regression from the image \cite{kanazawa2018endtoendrecoveryhumanshape}. Although robust, these methods capture only coarse geometry and lack photorealistic texture. Recent generative approaches have shown impressive progress in generating detailed geometry and appearance. Some methods use implicit representations such as NeRFs \cite{mildenhall2020nerfrepresentingscenesneural} to reconstruct the avatar \cite{hong2022eva3d, hong2022avatarclipzeroshottextdrivengeneration}. More recently, methods such as FaceLift \cite{lyu2024facelift} and GS-LRM \cite{zhang2024gslrm} have explored directly generating explicit representations such as 3D Gaussian splatting \cite{kerbl20233d}, achieving higher fidelity quality. However, all single-view methods are fundamentally limited by the available information, often leading to plausible but incorrect geometry for occluded regions (\eg, back view), which damages identity fidelity. Our work mitigates this by exploiting multiple unstructured input views.

\paragraph{Multi-view 3D human reconstruction.}
Using multiple views provides stronger geometric constraints for 3D reconstruction. Traditional MVS pipelines \cite{schoenberger2016sfm, schoenberger2016mvs} can produce high-quality results but require a large number of images with precise camera calibration. For humans, methods often rely on specialized capture setups, such as camera domes \cite{peng2021neuralbodyimplicitneural} or light stages. Recent works aim to relax these constraints, using neural representations to reconstruct avatars from sparse, calibrated video \cite{Lombardi_2019}. Our method takes this a step further, accepting a handful of uncalibrated and unstructured photos from a mobile device, using a generative model to bridge the gap to a structured multi-view format.

\paragraph{Generative models for 3D avatars.}
Generative models have become a cornerstone of 3D avatar creation. Diffusion models \cite{ho2020denoisingdiffusionprobabilisticmodels} and Transformers \cite{vaswani2017attentionneed} have been adapted to generate 3D assets, often conditioned on text or images. Large Reconstruction Models (LRMs) \cite{zhang2024gslrm, hong2024lrmlargereconstructionmodel} leverage transformers to predict a 3D representation from one or more input views. Our reconstruction model builds upon this paradigm. However, a key differentiator of our work is the data used for training. While most methods rely on synthetic datasets such as Objaverse \cite{deitke2022objaverseuniverseannotated3d} or RenderPeople, which lack fine details, we train on a novel dataset of high-fidelity 3D scans of real people. This allows our model to learn a much richer prior for realistic human appearance.
 
\section{Method}
\label{sec:method}

Our goal is to create a high-fidelity, identity-preserving 3D avatar from a few unstructured photos. We achieve this with our ``Capture, Canonicalize, Splat'' pipeline, illustrated in Figure~\ref{fig:pipeline}. The pipeline consists of two core components: a \textbf{generative canonicalization module} and a \textbf{multi-view 3D lifting module}. Crucially, both models are trained using our novel dataset of \textbf{person-specific Gaussian splatting avatars}, which is the key to achieve hyper-realism, leading to high reconstruction quality, as shown in Table~\ref{tab:ablation_table}.
\begin{figure}[ht]
  \centering
  \includegraphics{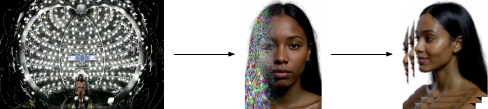}
  \caption{\textbf{Our high-fidelity dataset creation workflow.} (1) We start with calibrated multi-view images of real individuals from a dome capture setup. (2) We optimize a high-fidelity Gaussian avatar for each subject, which serves as our ground truth. (3) From this ground-truth 3D model, we render extensive training data, including canonical multi-view sets and simulated unstructured captures, to train our models.}
  \label{fig:psgs_flow}
\end{figure}

\subsection{A high-fidelity human avatar dataset}
\label{sec:dataset}

\begin{figure}[t]
  \centering
  \includegraphics{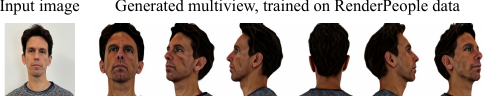} 
  \caption{\textbf{Limitations of synthetic training data.} A model trained solely on synthetic data such as RenderPeople fails to capture realism. Given a real input photo (left), its output (right) exhibits an identity shift and an overly smooth, stylized appearance.}
  \label{fig:synthetic_data_limitations}
\end{figure}

The quality of generative models is fundamentally tied to the quality of their training data. To overcome the realism gap of existing synthetic datasets, illustrated in Figure~\ref{fig:synthetic_data_limitations}, we created a new dataset of person-specific Gaussian splatting avatars. This dataset is derived from high-quality, multi-view dome captures of thousands of real individuals.

Our dataset creation workflow is illustrated in Figure~\ref{fig:psgs_flow}. For each subject, we begin by applying the universal avatar fitting pipeline of~\cite{li2024uravataruniversalrelightablegaussian}. This method takes as input a set of calibrated multi-view images captured in a controlled dome environment, and optimizes a 3D Gaussian splatting representation to match the subject's appearance and geometry. This process captures intricate, identity-specific details such as skin microgeometry, pores, fine wrinkles, and complex hair structures, which are often absent in standard synthetic assets. These initial GS avatars serve as our ground truth.
From these high-fidelity 3D models (3.2K avatars), we render large-scale (5M renders), multi-view datasets for training our components, including:
\begin{itemize}
    \item \textbf{Canonical multi-view sets:} To train our canonicalization model, we simulate realistic phone captures by rendering images with perturbations in camera position, orientation, and focal length, creating pairs of "unstructured inputs" and their corresponding "canonical ground truth".
    \item \textbf{Pose variation:} The pipeline supports rendering avatars in different body poses, further increasing the diversity and realism of the training data.
\end{itemize}
This data generation strategy enables training our models with a strong prior of realistic human appearance while maintaining precise 3D supervision. As shown in Table~\ref{tab:ablation_table}, training on our dataset yields substantial improvement over models trained on synthetic data like RenderPeople.
 
\subsection{Generative view canonicalization}
\label{sec:capturegen}

\begin{figure*}[t]
  \centering
  \includegraphics{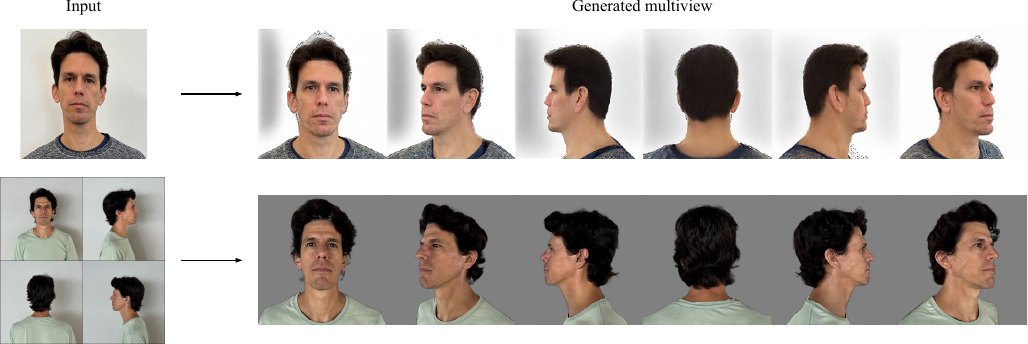} 
  \caption{\textbf{Importance of multiple views for identity preservation.} Reconstruction from a single view often fails to preserve identity, especially for unseen areas. Our multi-view approach produces a more faithful and consistent result.}
  \label{fig:multi_vs_single}
\end{figure*}

The first stage of our pipeline, the generative canonicalization module, is responsible for processing the input images. Given a small set of $N$ unstructured phone images $\{I_1, ..., I_N\}$ (typically $N=4$, corresponding to front, back, left, and right captures), the goal of this module is to produce a set of $M$ 3D-consistent, canonicalized views $\{C_1, ..., C_M\}$ with their corresponding fixed camera parameters $\{P_1, ..., P_M\}$. 
The canonicalization model is a foundation generative model designed for 3D-consistent image synthesis. It performs three key functions:
\begin{enumerate}
    \item \textbf{View normalization:} Aggregates identity information from all input views to produce a consistent appearance, effectively "averaging out" variations in lighting and pose.
    \item \textbf{3D consistency enforcement:} Trained to produce outputs that correspond to valid projections of a single underlying 3D object, ensuring geometric consistency across the canonical views.
    \item \textbf{Novel view synthesis:} Synthesizes views for which no direct input was provided (\eg, 45° views), filling in missing information to provide a comprehensive input for the subsequent 3D lifting stage.
\end{enumerate}
Our experiments show that the use of multiple input views is critical. As illustrated in Figure~\ref{fig:multi_vs_single}, a model conditioned on a single front view struggles to maintain identity in the side and back views. By conditioning on a sparse but holistic set of views, our canonicalization model significantly reduces hallucination and improves identity preservation.

\subsection{Multi-view to 3D Gaussian splatting reconstruction}
\label{sec:mv2splat}

The second stage, our multi-view reconstruction model, lifts the 2D canonical views into a 3D representation. It is a transformer-based Large Reconstruction Model, inspired by GS-LRM \cite{zhang2024gslrm}. The model takes as input the $M$ canonical images $\{C_j\}_{j=1}^M$ and their camera parameters $\{P_j\}_{j=1}^M$ generated by the canonicalization module.
The transformer architecture processes image features extracted from each view and predicts the properties of a set of $K$ 3D Gaussians, $\mathcal{G} = \{g_i\}_{i=1}^K$. Each Gaussian $g_i$ is defined by its properties: position $\boldsymbol{\mu}_i \in \mathbb{R}^3$, covariance $\boldsymbol{\Sigma}_i$ (represented as scale and rotation), color $\boldsymbol{c}_i \in \mathbb{R}^3$, and opacity $\alpha_i \in \mathbb{R}$.
The model is trained end-to-end on our high-fidelity avatar dataset. The training objective is a weighted sum of four components designed to ensure both photometric accuracy and geometric plausibility. First, we use a combination of an L1 photometric loss and a perceptual loss (LPIPS) \cite{zhang2018perceptual} to match the rendered color images with the ground truth. Second, we employ an alpha loss \cite{xu2024grm} that supervises the rendered alpha mask against the ground-truth foreground mask. This term is critical for removing floating artifacts and ensuring clean silhouettes. Finally, to prevent the formation of degenerate, needle-like Gaussians, we add a scale regularization loss that encourages more isotropic and compact Gaussians. The complete loss function is as follows:
\begin{equation}
  \mathcal{L} = \lambda_1 \mathcal{L}_{1} + \lambda_{p} \mathcal{L}_{\text{LPIPS}} + \lambda_{\alpha} \mathcal{L}_{\alpha} + \lambda_{\text{scale}} \mathcal{L}_{\text{scale}}
\end{equation}
Training directly on our high-fidelity data enables our reconstruction model to learn the intricate distributions of Gaussians required to represent fine details like hair and skin texture, which is the key to the hyper-realism of our final avatars. A critical finding is that our reconstruction model is highly sensitive to the 3D consistency of its inputs; the view normalization provided by the first stage is therefore essential for achieving high-quality, artifact-free reconstructions.
\begin{table}[!h]
  \centering
  \begin{tabular}{@{}lcc@{}}
    \toprule
    \textbf{Training data} & \textbf{Input views} & \textbf{PSNR \textuparrow} \\
    \midrule
    RenderPeople & Single & 25.3 \\
    RenderPeople & Multi-view & 27.5 \\
    \midrule
    Human Avatar Dataset & Single & 27.2 \\
    \textbf{Human Avatar Dataset} & \textbf{Multi-view} & \textbf{33.5} \\
    \bottomrule
  \end{tabular}
  \caption{\textbf{Ablation study on training data and input views.} Our full pipeline, trained on our high-fidelity Human Avatar Dataset with multi-view inputs, significantly outperforms other configurations on our internal test set. This demonstrates the critical importance of both high-quality training data and the use of multiple input views for high reconstruction quality.}
  \label{tab:ablation_table}
\end{table}

{
    \small
    \balance
    \bibliographystyle{ieeenat_fullname}
    \bibliography{main}

\begin{thebibliography}{20}
\providecommand{\natexlab}[1]{#1}
\providecommand{\url}[1]{\texttt{#1}}
\expandafter\ifx\csname urlstyle\endcsname\relax
  \providecommand{\doi}[1]{doi: #1}\else
  \providecommand{\doi}{doi: \begingroup \urlstyle{rm}\Url}\fi

\bibitem[Bogo et~al.(2016)Bogo, Kanazawa, Lassner, Gehler, Romero, and Black]{bogo2016smplautomaticestimation3d}
Federica Bogo, Angjoo Kanazawa, Christoph Lassner, Peter Gehler, Javier Romero, and Michael~J. Black.
\newblock Keep it {SMPL}: Automatic estimation of {3D} human pose and shape from a single image.
\newblock In \emph{ECCV, Part V}, pages 561--578, 2016.

\bibitem[Deitke et~al.(2023)Deitke, Schwenk, Salvador, Weihs, Michel, VanderBilt, Schmidt, Ehsani, Kembhavi, and Farhadi]{deitke2022objaverseuniverseannotated3d}
Matt Deitke, Dustin Schwenk, Jordi Salvador, Luca Weihs, Oscar Michel, Eli VanderBilt, Ludwig Schmidt, Kiana Ehsani, Aniruddha Kembhavi, and Ali Farhadi.
\newblock Objaverse: A universe of annotated {3D} objects.
\newblock In \emph{CVPR}, pages 13142--13153, 2023.

\bibitem[Ho et~al.(2020)Ho, Jain, and Abbeel]{ho2020denoisingdiffusionprobabilisticmodels}
Jonathan Ho, Ajay Jain, and Pieter Abbeel.
\newblock Denoising diffusion probabilistic models.
\newblock In \emph{NeurIPS}, pages 6840--6851, 2020.

\bibitem[Hong et~al.(2022)Hong, Zhang, Pan, Cai, Yang, and Liu]{hong2022avatarclipzeroshottextdrivengeneration}
Fangzhou Hong, Mingyuan Zhang, Liang Pan, Zhongang Cai, Lei Yang, and Ziwei Liu.
\newblock {AvatarCLIP}: Zero-shot text-driven generation and animation of {3D} avatars.
\newblock \emph{ACM TOG}, 41\penalty0 (4):\penalty0 161:1--161:19, 2022.

\bibitem[Hong et~al.(2023)Hong, Chen, Lan, Pan, and Liu]{hong2022eva3d}
Fangzhou Hong, Zhaoxi Chen, Yushi Lan, Liang Pan, and Ziwei Liu.
\newblock {EVA3D}: Compositional {3D} human generation from {2D} image collections.
\newblock In \emph{ICLR}, 2023.

\bibitem[Hong et~al.(2024)Hong, Zhang, Gu, Bi, Zhou, Liu, Liu, Sunkavalli, Bui, and Tan]{hong2024lrmlargereconstructionmodel}
Yicong Hong, Kai Zhang, Jiuxiang Gu, Sai Bi, Yang Zhou, Difan Liu, Feng Liu, Kalyan Sunkavalli, Trung Bui, and Hao Tan.
\newblock {LRM}: Large reconstruction model for single image to {3D}.
\newblock In \emph{ICLR}, 2024.

\bibitem[Kanazawa et~al.(2018)Kanazawa, Black, Jacobs, and Malik]{kanazawa2018endtoendrecoveryhumanshape}
Angjoo Kanazawa, Michael~J. Black, David~W. Jacobs, and Jitendra Malik.
\newblock End-to-end recovery of human shape and pose.
\newblock In \emph{CVPR}, pages 7122--7131, 2018.

\bibitem[Kerbl et~al.(2023)Kerbl, Kopanas, Leimkühler, and Drettakis]{kerbl20233d}
Bernhard Kerbl, Georgios Kopanas, Thomas Leimkühler, and George Drettakis.
\newblock {3D} {Gaussian} splatting for real-time radiance field rendering.
\newblock \emph{ACM TOG}, 42\penalty0 (4):\penalty0 139:1--139:14, 2023.

\bibitem[Li et~al.(2024)Li, Cao, Schwartz, Khirodkar, Richardt, Simon, Sheikh, and Saito]{li2024uravataruniversalrelightablegaussian}
Junxuan Li, Chen Cao, Gabriel Schwartz, Rawal Khirodkar, Christian Richardt, Tomas Simon, Yaser Sheikh, and Shunsuke Saito.
\newblock {URAvatar}: Universal relightable {Gaussian} codec avatars.
\newblock In \emph{SIGGRAPH Asia 2024 Conference Papers}, pages 128:1--128:11, 2024.

\bibitem[Lombardi et~al.(2019)Lombardi, Simon, Saragih, Schwartz, Lehrmann, and Sheikh]{Lombardi_2019}
Stephen Lombardi, Tomas Simon, Jason Saragih, Gabriel Schwartz, Andreas Lehrmann, and Yaser Sheikh.
\newblock Neural volumes: Learning dynamic renderable volumes from images.
\newblock \emph{ACM TOG}, 38\penalty0 (4):\penalty0 65:1--65:14, 2019.

\bibitem[Loper et~al.(2015)Loper, Mahmood, Romero, Pons-Moll, and Black]{smpl}
Matthew Loper, Naureen Mahmood, Javier Romero, Gerard Pons-Moll, and Michael~J. Black.
\newblock {SMPL}: A skinned multi-person linear model.
\newblock \emph{ACM TOG}, 34\penalty0 (6):\penalty0 248:1--248:16, 2015.

\bibitem[Lyu et~al.(2025)Lyu, Zhou, Yang, and Shu]{lyu2024facelift}
Weijie Lyu, Yi Zhou, Ming-Hsuan Yang, and Zhixin Shu.
\newblock {FaceLift}: Learning generalizable single image {3D} face reconstruction from synthetic heads.
\newblock In \emph{ICCV}, 2025.
\newblock To appear.

\bibitem[Mildenhall et~al.(2020)Mildenhall, Srinivasan, Tancik, Barron, Ramamoorthi, and Ng]{mildenhall2020nerfrepresentingscenesneural}
Ben Mildenhall, Pratul~P. Srinivasan, Matthew Tancik, Jonathan~T. Barron, Ravi Ramamoorthi, and Ren Ng.
\newblock {NeRF}: Representing scenes as neural radiance fields for view synthesis.
\newblock In \emph{ECCV, Part I}, pages 405--421, 2020.

\bibitem[Peng et~al.(2021)Peng, Zhang, Xu, Wang, Shuai, Bao, and Zhou]{peng2021neuralbodyimplicitneural}
Sida Peng, Yuanqing Zhang, Yinghao Xu, Qianqian Wang, Qing Shuai, Hujun Bao, and Xiaowei Zhou.
\newblock Neural body: Implicit neural representations with structured latent codes for novel view synthesis of dynamic humans.
\newblock In \emph{CVPR}, pages 9050--9059, 2021.

\bibitem[Schönberger and Frahm(2016)]{schoenberger2016sfm}
Johannes~L. Schönberger and Jan-Michael Frahm.
\newblock Structure-from-motion revisited.
\newblock In \emph{CVPR}, pages 4104--4113, 2016.

\bibitem[Schönberger et~al.(2016)Schönberger, Zheng, Pollefeys, and Frahm]{schoenberger2016mvs}
Johannes~L. Schönberger, Enliang Zheng, Marc Pollefeys, and Jan-Michael Frahm.
\newblock Pixelwise view selection for unstructured multi-view stereo.
\newblock In \emph{ECCV, Part III}, pages 501--518, 2016.

\bibitem[Vaswani et~al.(2017)Vaswani, Shazeer, Parmar, Uszkoreit, Jones, Gomez, Kaiser, and Polosukhin]{vaswani2017attentionneed}
Ashish Vaswani, Noam Shazeer, Niki Parmar, Jakob Uszkoreit, Llion Jones, Aidan~N Gomez, \L{}ukasz Kaiser, and Illia Polosukhin.
\newblock Attention is all you need.
\newblock In \emph{NIPS}, pages 5999--6009, 2017.

\bibitem[Xu et~al.(2024)Xu, Shi, Yifan, Chen, Yang, Peng, Shen, and Wetzstein]{xu2024grm}
Yinghao Xu, Zifan Shi, Wang Yifan, Hansheng Chen, Ceyuan Yang, Sida Peng, Yujun Shen, and Gordon Wetzstein.
\newblock {GRM}: Large {Gaussian} reconstruction model for efficient {3D} reconstruction and generation.
\newblock In \emph{ECCV, Part XV}, pages 1--20, 2024.

\bibitem[Zhang et~al.(2024)Zhang, Bi, Tan, Xiangli, Zhao, Sunkavalli, and Xu]{zhang2024gslrm}
Kai Zhang, Sai Bi, Hao Tan, Yuanbo Xiangli, Nanxuan Zhao, Kalyan Sunkavalli, and Zexiang Xu.
\newblock {GS-LRM}: Large reconstruction model for {3D} {Gaussian} splatting.
\newblock In \emph{ECCV, Part XXII}, pages 1--19, 2024.

\bibitem[Zhang et~al.(2018)Zhang, Isola, Efros, Shechtman, and Wang]{zhang2018perceptual}
Richard Zhang, Phillip Isola, Alexei~A. Efros, Eli Shechtman, and Oliver Wang.
\newblock The unreasonable effectiveness of deep features as a perceptual metric.
\newblock In \emph{CVPR}, pages 586--595, 2018.

\end{thebibliography}
}
\end{document}